# COMPARATIVE STUDY OF IMAGE EDGE DETECTION ALGORITHMS

## Shubham Saini[1], Bhavesh Kasliwal[2], Shraey Bhatia[3]

[1] *Student, School of Computing Science and Engineering, Vellore Institute of Technology, India,*
*shubham.saini2010@vit.ac.in*

[2] *Student, School of Computing Science and Engineering, Vellore Institute of Technology, India,*
*bhavesh.kasliwal2010@vit.ac.in*

[3] *Student, School of Computing Science and Engineering, Vellore Institute of Technology, India,*
*shraey.bhatia2010@vit.ac.in*

### *Abstract*

*Since edge detection is in the forefront of image processing for object detection, it is crucial to have a good understanding of edge detection algorithms. The reason for this is that edges form the outline of an object. An edge is the boundary between an object and the background, and indicates the boundary between overlapping objects. This means that if the edges in an image can be identified accurately, all of the objects can be located and basic properties such as area, perimeter, and shape can be measured. Since computer vision involves the identification and classification of objects in an image, edge detections is an essential tool. We tested two edge detectors that use different methods for detecting edges and compared their results under a variety of situations to determine which detector was preferable under different sets of conditions.*

## I.    Introduction

Edge detection is a very important area in the field of Computer Vision. Edges define the boundaries between regions in an image, which helps with segmentation and object recognition. They can show where shadows fall in an image or any other distinct change in the intensity of an image. Edge detection is a fundamental of low-level image processing and good edges are necessary for higher level processing. [1]

Since different edge detectors work better under different conditions, it would be ideal to have an algorithm that makes use of multiple edge detectors, applying each one when the scene conditions are most ideal for its method of detection. In order to create this system, you must first know which edge detectors perform better under which conditions. That is the goal of our project. We tested two edge detectors that use different methods for detecting edges and compared their results under a variety of situations to determine which detector was preferable under different sets of conditions.

## II.    Classification of Edge Detectors

A. **Laplacian of Gaussian (LoG):** Which was invented by Marr and Hildreth (1980) who combined Gaussian filtering with the Laplacian. This algorithm is not used frequently in machine vision. Those who continued his way were Berzins (1984), Shah, Sood and Jain (1986), Huertas and Medioni (1986) [2,3].

B. **Gaussian Edge Detectors:** This is symmetric along the edge and reduces the noise by smoothing the image. The significant operators here are Canny and ISEF (Shen-Castan) which convolve the image with the derivative of Gaussian for Canny and ISEF for Shen- Castan.[2,3].

## III.    The Marr-Hildreth Edge Detector

The Marr-Hildreth edge detector was a very popular edge operator before Canny released his paper. It is a gradient based operator which uses the Laplacian to take the second derivative of an image. The idea is that if there is a step difference in the intensity of the image, it will be represented by in the second derivative by a zero crossing.

So the general algorithm for the Marr-Hildreth edge detector is as follows:

**1.** Smooth the image using a Gaussian. This smoothing reduces the amount of error found due to noise.

**2.** Apply a two dimensional Laplacian to the image:

This Laplacian will be rotation invariant and is often called the "Mexican Hat operator" because of its shape:

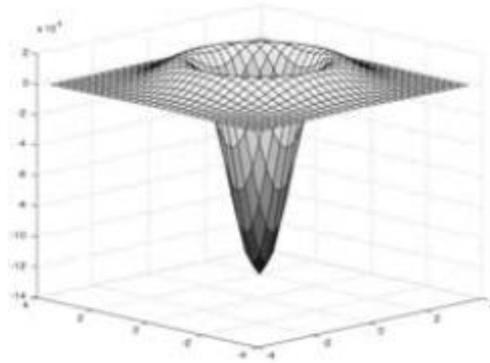

This operation is the equivalent of taking the second derivative of the image.
**3.** Loop through every pixel in the Laplacian of the smoothed image and look for sign changes. If there is a sign change and the slope across this sign change is greater than some threshold, mark this pixel as an edge. Alternatively, you can run these changes in slope through a hysteresis (described in the canny edge detector) rather than using a simple threshold.
(Algorithm taken from [4])

## IV.     The Canny Edge Detector

The Canny edge detector is widely considered to be the standard edge detection algorithm in the industry. It was first created by John Canny for his Master's thesis at MIT in 1983 [5], and still outperforms many of the newer algorithms that have been developed. Canny saw the edge detection problem as a signal processing optimization problem, so he developed an objective function to be optimized [5]. The solution to this problem was a rather complex exponential function, but Canny found several ways to approximate and optimize the edge-searching problem. The steps in the canny edge detector are as follows:
**1.** Smooth the image with a two dimensional Gaussian. In most cases the computation of a two dimensional Gaussian is costly, so it is approximated by two one dimensional Gaussians, one in the x direction and the other in the y direction.
**2.** Take the gradient of the image. This shows changes in intensity, which indicates the presence of edges. This actually gives two results, the gradient in the x direction and the gradient in the y direction.
**3.** Non-maximal suppression. Edges will occur at points the where the gradient is at a maximum. Therefore, all points not at a maximum should be suppressed. In order to do this, the magnitude and direction of the gradient is computed at each pixel. Then for each pixel check if the magnitude of the gradient is greater at one pixel's distance away in either the positive or the negative direction perpendicular to the gradient. If the pixel is not greater than both, suppress it.

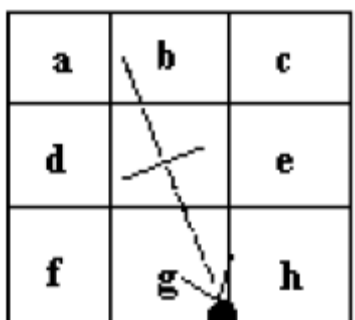

**4.** Edge Thresholding. The method of thresholding used by the Canny Edge Detector is referred to as "hysteresis". It makes use of both a high threshold and a low threshold. If a pixel has a value above the high threshold, it is set as an edge pixel. If a pixel has a value above the low threshold and is the neighbor of an edge pixel, it is set as an edge pixel as well. If a pixel has a value above the low threshold but is not the neighbor of an edge pixel, it is not set as an edge pixel. If a pixel has a value below the low threshold, it is never set as an edge pixel.
(Algorithm based on description given in [6])

## V. Comparison of Edge Detectors

In the past two decades several algorithms have been developed to extract edges within digital images but their functionalities and performances are not the same. In spite of all the efforts, none of the proposed operators are fully satisfactory in real world. The availability of well-defined quality criteria is important.

There are five different criteria that are typically used for testing the quality of an edge detector:
-The probability of a false positive (marking something as an edge which isn't an edge)
-The probability of a false negative (failing to mark an edge which actually exists)
-The error in estimating the edge angle
-The mean square distance of the edge estimate from the true edge
-The algorithm's tolerance to distorted edges and features such as corners and junctions
(Criteria taken from [5])

However, in order to determine the third and fourth criteria, an exact map of the edges in an image must be known, and in general this is not available. It is also not plausible to assume that some "exact map" of all the edges can even be constructed. Therefore, the third and fourth criteria are not very useful. Additionally, corners and junctions simply are not handled well by any edge detector and must be considered separately. Therefore the fifth criterion is not very useful either.

The most important criteria are the first two, as it is much more important to have the proper features labeled as edges than having each edge exactly follow what would be considered the "ideal" edge or being able to handle special features such as corners and junctions.

## VI. EXPERIMENTAL RESULTS

All color images were converted to grayscale using MATLAB's RBG2GRAY function.

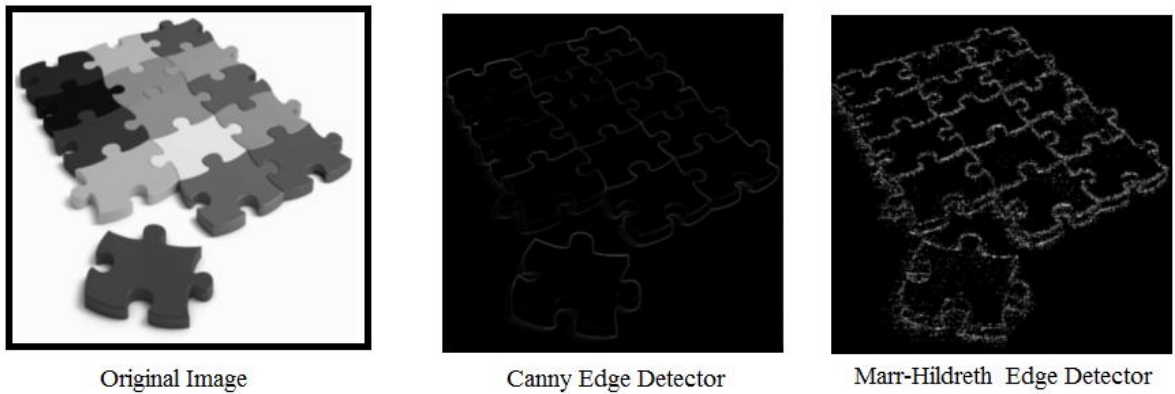

Original Image     Canny Edge Detector     Marr-Hildreth Edge Detector

FIGURE 1

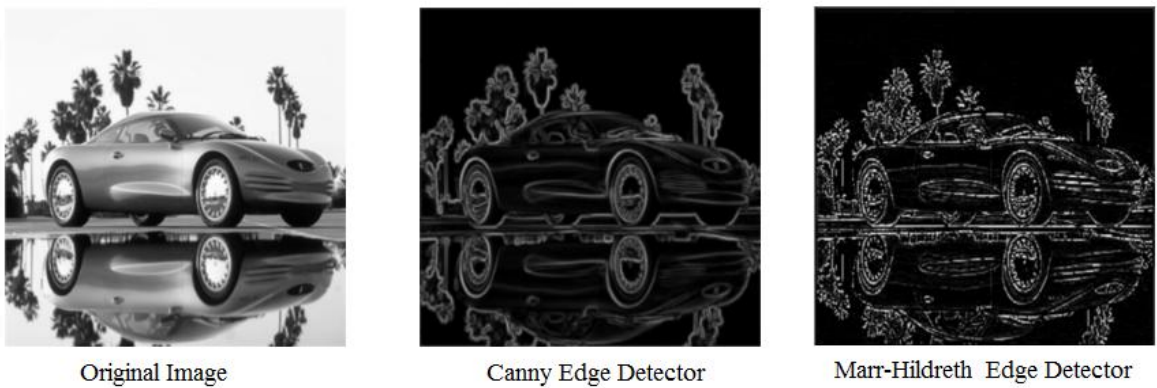

Original Image     Canny Edge Detector     Marr-Hildreth Edge Detector

FIGURE 2

The visual comparison of the above two sets of images (in Figures 1 and 2) can lead us to the subjective evaluation of the performances of selected edge detectors. Applying these two methods to a noisy image shows that with noisy images, second derivative operators, like Canny, exhibit better performance but require more computations because of smoothing an image with a Gaussian function first and then computing the gradient.

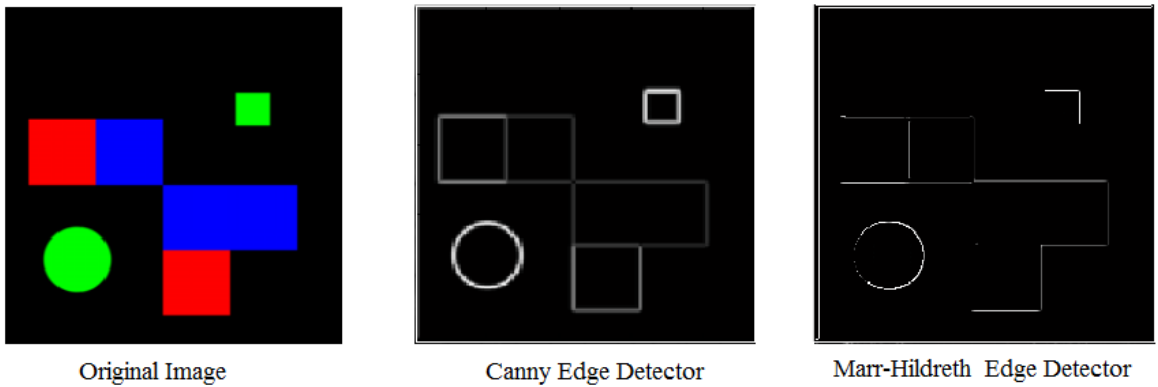

FIGURE 3

Figure 3 shows the ability of the edge detectors to handle corners as well as a wide range of slopes in edge on the circle. The Canny edge detector becomes fairly confused at corners due to the Gaussian smoothing of the image. Also, since the direction of the edge changes instantly, corner pixels looks in the wrong directions for its neighbors. The Marr-Hildreth edge detector fails to detect many prominent edges.

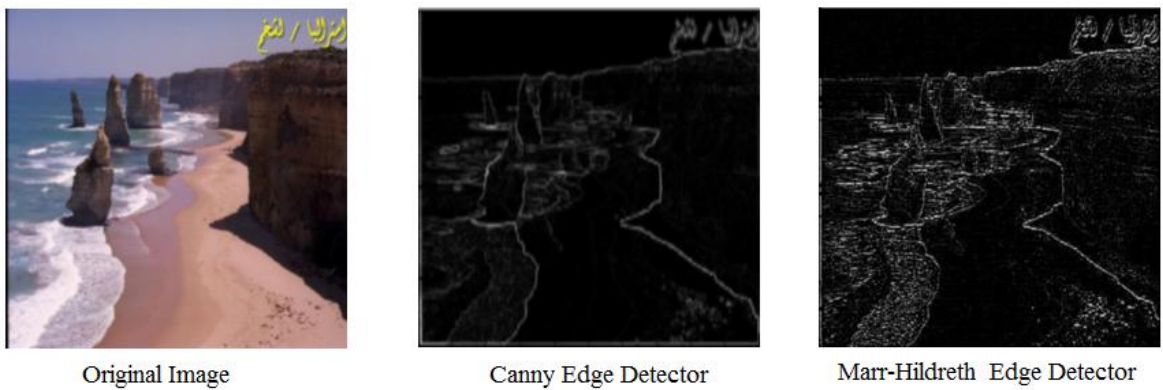

FIGURE 4

Figure 4 is a picture of a shoreline. Both the edge detectors had problems detecting the different ridges of the cliff. The foam of the waves also provided some inconsistent results. There are a lot of discrepancies in color at these locations, but no clear edges.

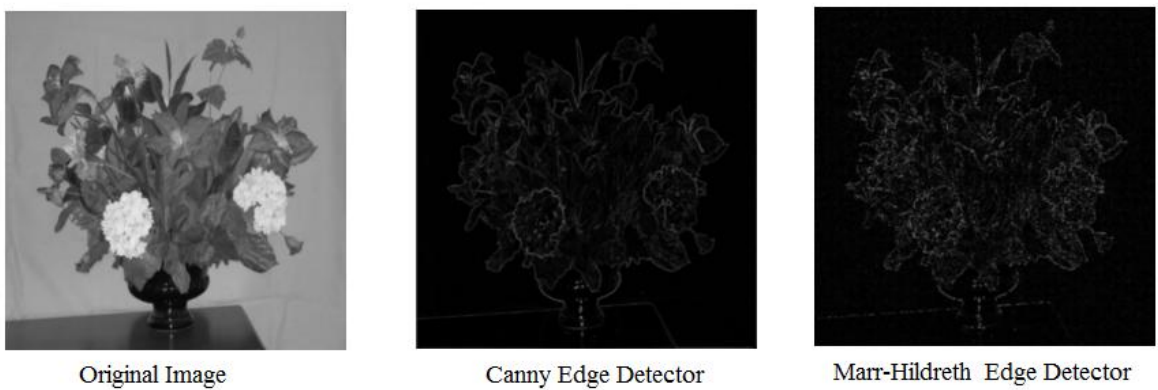

FIGURE 5

Figure 5 more accurately shows the capabilities of each edge detector. The Marr-Hildreth detector perceives many edges, but they are too spotty and wide to really identify any features. The Canny edge detector gives nice outlines of the table, the vase, and many of the flowers on the border. Features in the middle of the arrangement are missed, but some are recovered with the addition of color. For example, the red flower in the center and the leaves to its right are found.

Canny has specified three issues that an edge detector must address in order to better detect edges in noise conditions: Error rate, Localization and Response.

In Marr-Hildreth, locality is not especially good and the edges are not always thin, still this edge detector is much better than the classical ones in cases of low signal to noise ratio. The Marr-Hildreth edge detector will give more nicely connected edges if hysteresis is used to threshold the image, and will not give connected edges if a single threshold is used. Regardless, it is usually gives spotty and thick edges.

## CONCLUSION

Since edge detection is the initial step in object recognition, it is necessary to know the differences between edge detection algorithms. In Marr-Hildreth, locality is not especially good and the edges are not always thin. Canny's method is preferred since it produces single pixel thick, continuous edges.